\documentclass{article}

\usepackage{graphicx}
\usepackage{float}
\usepackage{caption}
\usepackage{subcaption} 
\usepackage[super]{nth}
\usepackage{amsmath}
\usepackage{bm}
\usepackage{hyperref}
\usepackage[dvipsnames]{xcolor}  %color
\usepackage{lineno} %line number
\usepackage{placeins}
\usepackage{array}
\usepackage{tabu}
\usepackage{amssymb}
\usepackage{mathtools}
\usepackage{lineno}
\usepackage{wrapfig}
\usepackage{booktabs}       % professional-quality tables
\usepackage{float}
\usepackage{authblk}
\usepackage{enumerate}
\usepackage{xcolor}
\usepackage{multirow}
\usepackage{enumitem}
\usepackage{amsthm}
\usepackage{amsmath}
\usepackage{comment}
\usepackage{booktabs} % for professional tables
\usepackage{pdflscape}
\usepackage{listings}

\def\R{\mathbb{R}}

\usepackage[accepted]{icml2024}

\usepackage[capitalize,noabbrev]{cleveref}

\theoremstyle{plain}

\theoremstyle{definition}

\theoremstyle{remark}

\usepackage[textsize=tiny]{todonotes}

\icmltitlerunning{Fine-Tune Language Models as Multi-Modal Differential Equation Solvers}

\begin{document}

\twocolumn[
\icmltitle{Fine-Tune Language Models as Multi-Modal Differential Equation Solvers}




\begin{icmlauthorlist}
\icmlauthor{Liu Yang}{ucla}
\icmlauthor{Siting Liu}{ucla}
\icmlauthor{Stanley J. Osher}{ucla}
\end{icmlauthorlist}

\icmlaffiliation{ucla}{Department of Mathematics, University of California, Los Angeles, Los Angeles, CA, USA}

\icmlcorrespondingauthor{Stanley J. Osher}{sjo@math.ucla.edu}

\icmlkeywords{Machine Learning, ICML}

\vskip 0.3in
]



\printAffiliationsAndNotice{}  % leave blank if no need to mention equal contribution

\begin{abstract}
In the growing domain of scientific machine learning, in-context operator learning has shown notable potential in building foundation models, as in this framework the model is trained to learn operators and solve differential equations using prompted data, during the inference stage without weight updates. However, the current model's overdependence on function data overlooks the invaluable human insight into the operator. To address this, we present a transformation of in-context operator learning into a multi-modal paradigm. In particular, we take inspiration from the recent success of large language models, and propose using ``captions'' to integrate human knowledge about the operator, expressed through natural language descriptions and equations. Also, we introduce a novel approach to train a language-model-like architecture, or directly fine-tune existing language models, for in-context operator learning. We beat the baseline on single-modal learning tasks, and also demonstrated the effectiveness of multi-modal learning in enhancing performance and reducing function data requirements. The proposed method not only significantly enhanced the development of the in-context operator learning paradigm, but also created a new path for the application of language models.

\end{abstract}

\section{Introduction}

\begin{figure*}[ht]
\begin{center}
\includegraphics[width=\textwidth]{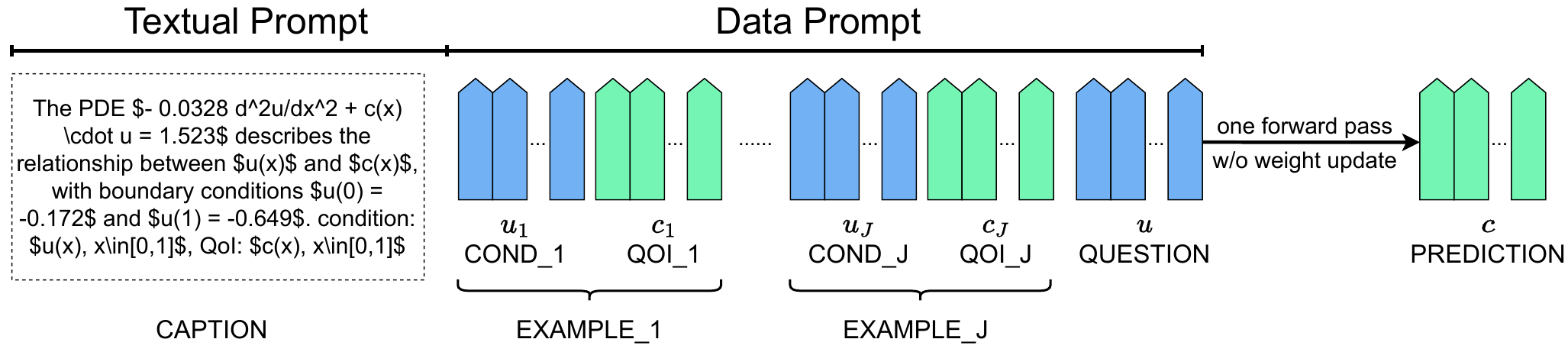}
% \vskip -0.1in
\caption{Diagram for multi-modal in-context operator learning.}
\label{fig:multi-modal}
\end{center}
\vskip -0.2in
\end{figure*}

Recently, in-context operator learning and the corresponding model In-Context Operator Networks (ICON) \cite{yang2023context} has been proposed as a new paradigm for scientific machine learning. 

As in classic operator learning tasks, an operator maps a single input function or a tuple of input functions, referred to as the ``condition'', to an output function, referred to as the ``quantity of interest (QoI)''. In practice, we usually have no access to the analytical expression of these functions, but instead can collect function data in the form of key-value pairs, where the keys are discrete function inputs and the values are the corresponding function outputs.

A wide variety of scientific machine learning tasks can be conceptualized as operator learning problems. Consider the task of solving partial differential equations (PDEs) for instance, where the coefficient function serves as the condition, and the solution is the QoI. Conversely, for inverse problems, these roles are swapped. When dealing with problems involving temporal evolution, the condition can be the initial function, while the QoI represents the function at a later time. For control problems, the condition could correspond to the cost function and the initial state, while the QoI embodies the control signal. It's evident that the relationship between the condition and the QoI highly depends on the operator, which is defined by the task at hand and the particular system in question.

In classic operator learning approaches~\cite{chen1995universal, chen1995approximation,khoo2021solving,zhu2018bayesian,long2018pde,lu2021learning,wang2021learning,li2021fourier,kovachki2023neural,bhattacharya2021model,li2021physics,subramanian2023towards}, a neural network is limited to approximate a specific operator, and thus need to be trained every time a new operator is encountered. In contrast, in-context operator learning aims to train the model as an ``operator learner'' instead of an ``operator approximator''. In particular, the model is trained to learn the operator from the prompted condition-QoI pairs, referred to as ``examples'', and apply the learned operator to the question condition to predict the corresponding QoI. After training, the above learning process can be performed through one forward pass of the model, in the inference stage without weight update. This approach offers a ``train-once-apply-multiple'' paradigm and paves the way for large-scale foundation models~\cite{bommasani2021opportunities} for a broad array of scientific machine learning tasks.

The study of ICON showcases the successful implementation of in-context operator learning, which relies solely on function data. However, a crucial aspect of scientific machine learning is overlooked in this approach, namely, the human knowledge of the operator, which can span from vague natural language explanations to explicit differential equations. There's a strong case for incorporating such knowledge into the learning system alongside function data, as this could potentially enhance learning performance. If the human understanding of the operator is sufficiently detailed, the system might require fewer examples to learn the operator. Theoretically, it might even enable zero-shot learning, where the operator could be learned and utilized without the need for any examples.

Past research on the topic of scientific machine learning typically integrates human knowledge into the learning system by designing special loss functions or neural network architectures based on the differential equations or symmetry/conservation laws that govern the system. While these approaches have witnessed significant success, they are not without limitations. Firstly, it may not always be practical to design special loss functions or architectures, as the system might not be fully understood by humans, or the operator might be too complicated to be described by equations. Secondly, these bespoke loss functions or architectures are tailored for specific systems or tasks. When confronted with a new system, there is a requirement not only to design new loss functions or architectures but also typically to retrain the neural network.

In this paper, we explore an entirely different approach to infusing human knowledge into the learning system. Inspired by the recent success of large language models (LLMs), we introduce a new component to in-context operator learning: the ``caption''. A caption is a string serving as a descriptor of the operator, and can take various forms such as equations written in LaTeX forms, natural language descriptions, or a combination of both. Rather than crafting special loss functions or architectures, we simply feed the caption into the neural network as input alongside the examples. We thus evolve the in-context operator learning to be multi-modal, meaning that the neural network can learn the operator from function data, captions, or a combination of both, as illustrated in Figure \ref{fig:multi-modal}.

We also introduce a novel approach to train a language-model-like architecture, or directly fine-tune existing language models, for single-modal and multi-modal in-context operator learning. The improved training scheme mimics the ``next-token prediction'' in LLMs: the model predicts the QoI in each example based on previous examples. We call it ``next-function prediction''. The main deviation (and also the key challenge) is the necessity to design the input sequence and formulate a specialized mask to accommodate in-context operator learning tasks. Following the name ``In-Context Operator Networks (ICON)'', we refer to our architecture and training scheme as ``ICON-LM'', where ``LM'' stands for ``language model''.

The adoption of language models for in-context learning is crucial for two reasons. First, it enables us to utilize existing ecosystem developed for language models for in-context operator learning. Second, it paves the way to broaden the capability of language models to scientific machine learning tasks with heavy numerical computations.

Our contributions are summarized as follows:
\begin{enumerate}
    \item We transform the in-context operator learning into a multi-modal framework by introducing ``captions'' as a means to incorporate human knowledge about the operator, in the form of natural language descriptions and equations.
    \item We introduce a novel approach, namely ``ICON-LM'', to train a language-model-like architecture, or directly fine-tune existing language models, for in-context operator learning. We beat the baseline on single-modal learning tasks, and also demonstrated the effectiveness of multi-modality with ICON-LM in enhancing performance and reducing function data requirements.
    \item By bridging language models with data-driven differential equation solvers, we have not only achieved substantial advancements in this specific domain of operator learning, but also opened up a new avenue for the application of language models in scientific machine learning, especially in areas that require heavy numerical computations.
\end{enumerate}

The rest of the paper is organized as follows. In \cref{sec:work}, we review the related work. We introduce the dataset in \cref{sec:data}. In \cref{sec:method}, we introduce the ICON-LM architecture and training scheme. In \cref{sec:exp}, we present the experimental results. We conclude in \cref{sec:sum}.

\section{Related Work}\label{sec:work}

\subsection{Operator Learning and In-Context Operator Learning}
Numerous neural network methods have been proposed for approximating operators, i.e., mappings that take functions as input and output.  The early works of~\cite{chen1995universal, chen1995approximation} employed shallow neural networks for the approximation of nonlinear operators. A deep neural network approach to tackle parametric PDE challenges was suggested in~\cite{khoo2021solving}. PDE-Net, as presented in~\cite{long2018pde} enables forward predictions of PDE solutions using the inferred forward map. The study in~\cite{zhu2018bayesian} presented a Bayesian method to address uncertainty quantification in stochastic PDE scenarios. The Deep Operator Network (DeepONet), referenced in~\cite{lu2021learning}, introduces a neural network design that approximates the solution operator, mapping parameters or initial/boundary conditions to their corresponding solutions. The Fourier Neural Operator (FNO) from~\cite{li2021fourier,kovachki2023neural} leverages the Fourier space's integral kernel to approximate the solution operator. Drawing inspiration from neural networks and model reduction, the paper~\cite{bhattacharya2021model} estimates input-output maps between infinite-dimensional spaces for parametric PDEs. Additional contributions can be found in~\cite{kochkov2021machine,kissas2022learning,goswami2022deep,zhu2023reliable,subel2023explaining}.

Recently, a different paradigm, namely in-context operator learning, is proposed in~\cite{yang2023context}, which is an extension of in-context learning introduced in GPT-2~\cite{radford2019language} and GPT-3~\cite{brown2020language}.
Instead of approximating specific operators, in-context operator learning trains the neural network as an operator learner, which can learn and apply the operator through one forward pass of the model, in the inference stage without weight update. Such in-context learning capability can even generalize to new equations~\cite{liu2023does,yang2024pde}.

\subsection{Physics-Informed Machine Learning}

In the literature, two approaches are commonly employed to incorporate physical knowledge in neural networks: hard constraints and soft constraints. We refer readers to the survey paper~\cite{karniadakis2021physics} on this topic. Hard constraints involve designing neural network architectures in a way that ensures any solution generated by the network meets the specified constraints, for example,~\cite{zhang2018deep,pfau2020ab,pun2019physically,ling2016reynolds,jin2020sympnets,lusch2018deep,mattheakis2019physical}.
While solutions with specifically designed architectures are guaranteed to be compliant to the physical constraints, creating such architectures demands extensive domain knowledge and may not be easily adaptable to other problems. Additionally, the expressivity and training complexity could be limited in these cases.
Soft constraints are implemented by incorporating physics-informed terms into the loss function. For example, ~\cite{han2017deep, Han2018Solving, Sirignano2018DGM, E2018deepRitz, Raissi2019PINN, Zang2020Weakadversarial, Ruthotto2020machine,li2021physics,wang2021learning}. While more flexible in terms of neural network architecture design, this approach still requires precise knowledge of physics in the form of differential equations, variational problems, etc., which are not always available, especially when the system is not fully understood by humans.  

In-context operator learning excels at addressing a broad range of physical problems using a single neural network. The limited flexibility and generalizability of the previously mentioned approaches hinder their application to in-context operator learning. This limitation motivates our exploration in this paper, where we introduce a new method to incorporate physical knowledge: through ``captions''.

\subsection{Multi-Modal Models}

Unimodal language models solely rely on text data for training, limiting their ability to comprehend the visual world. In contrast, multimodal language models are trained on data in multiple forms, including texts and images, enabling them to understand the visual world. We refer readers to the survey~\cite{yin2023survey} on this topic.

To fuse different modal data, one approach involves combining the extracted features or embeddings from different modal data and then feeding these embeddings into the same model \cite{driess2023palm,alayrac2022flamingo,li2023blip,zhang2023pmc,liu2023visual,zhang2023llama,pi2023detgpt,zhang2023video,tsimpoukelli2021multimodal,zhu2023minigpt,rt22023arxiv}. Another approach converts other modal data into language data and uses these language representations as inputs for language models \cite{yang2022empirical}. Some studies combine both techniques, utilizing both extracted features and converted language data as inputs to language models \cite{li2023videochat,gao2023llama}.

During the training phase, due to the constraints of computational complexity, many models (such as \cite{tsimpoukelli2021multimodal,zhu2023minigpt,alayrac2022flamingo}) freeze the parameters of the language models and only train the parameters of the other components, like data embeddings or bridging projections. The performance of this training strategy is compared to end-to-end fine-tuning in \cite{driess2023palm}.

In this paper, we integrate embeddings of function data and language captions into the same model, instead of converting one form of data into another, and fine-tune the model in an end-to-end way.

\section{Dataset}\label{sec:data}

In this study, we inherit the function data from \cite{yang2023context}. This dataset contains 19 types of operator learning problems, including forward and inverse ordinary differential equations (ODEs), partial differential equations (PDEs), and mean-field control problems, with each type containing infinite operators. We list the 19 types of problems and an illustration in \cref{sec:ap_problem}.

Within the training dataset, each problem type comes with 1,000 distinct operators characterized by hidden parameters. For every operator, there are 100 condition-QoI pairs governed by such a shared operator. During training, one can randomly sample from these 100 pairs as ``examples'' and ``questions'' to build an instance of prompt and label. In the testing dataset, each problem type is represented by an additional 100 unique operators. Every operator is associated with 5 sets of condition-QoI pairs, and each set has 6 such pairs. For testing purposes, the initial $J$ pairs in each set can serve as ``examples'', while the final pair acts as the ``question'' for $J$-shot learning, with $J$ ranging from zero to five. This means the testing dataset consists of $19\times500$ sets, translating to $19\times500$ learning cases for every value of $J$.

In these problems, the condition/QoI function is in a 1D or 2D domain, depending on the problem type. The dataset stores the function on grids. We can sample data points from the grid for each function to construct data prompts that represent the function.

For multi-modal learning, we produced 160 captions per operator for training and an additional 40 for testing. These captions are evenly split into two categories: vague and precise, depending on whether they reveal the actual parameter values that determine the operator, e.g., the decay rate of a damped oscillator, the boundary condition of PDEs, or the terminal cost in a mean-field control problem. 

These caption data are generated with the assistance of GPT-4. We provided the details of caption data generation in \cref{sec:ap_caption_gen}. All the captions are open-sourced alongside code, with some examples listed in \cref{sec:ap_caption_ex}.
\section{ICON-LM Model}\label{sec:method}

\begin{table*}[!htp]
\caption{The tokens for the $j$-th example for the one-dimensional forward ODE problem.}
\vskip 0.15in
\centering
{
\footnotesize
\begin{tabular}{ll|l|l|l}
\toprule
&\multicolumn{1}{c|}{} & \multicolumn{1}{c|}{condition} & \multicolumn{1}{c|}{QoI} & \multicolumn{1}{c}{query} \\
\midrule
\multicolumn{1}{c|}{}  &term &\multirow{4}{*}{$\begin{pmatrix} 0 & 0 & \dots & 0 & 1\\ t_1 & t_2 & \dots & t_{n_j-1} & 0\\ 0 & 0 & \dots & 0 & 0\\ c(t_1) & c(t_2) & \dots & c(t_{n_j-1}) & u(0) \end{pmatrix}$ } &\multirow{4}{*}{$\begin{pmatrix} 0 & 0 & \dots & 0\\ \tau_1 & \tau_2 & \dots & \tau_{m_j}\\ 0 & 0 & \dots & 0\\ u(\tau_1) & u(\tau_2) & \dots & u(\tau_{m_j}) \end{pmatrix}$} &\multirow{4}{*}{$\begin{pmatrix} 0 & 0 & \dots & 0\\ \tau_1 & \tau_2 & \dots & \tau_{m_j}\\ 0 & 0 & \dots & 0\\ 0 & 0 & \dots & 0\end{pmatrix}$} \\
\multicolumn{1}{c|}{key} &time & & & \\
\multicolumn{1}{c|}{} &space & & & \\
\cline{1-2}
\multicolumn{2}{c|}{value} & & & \\
\bottomrule
\end{tabular}
}
\label{tab:sec2_demok_linearode}
\vskip -0.1in
\end{table*}

\subsection{Overview}

The input prompt of ICON-LM model consists of two parts: captions and function data. The data prompt consists of conditions and QoIs, each represented by a set of tokens. Each token is constructed by a key-value pair, where conceptually the key is the function input, including temporal and spatial coordinates, and the value is the corresponding function outputs. Sometimes a condition/QoI consists of multiple terms. For example, in forward ODE problems, the condition consists of the control function as well as the initial condition. To handle such scenarios, we also add a ``term'' indicator to the key to distinguish multiple terms.

The original ICON architecture consists of two transformers: an encoder and a decoder. For every prompt instance, the model is solely trained to perform in-context operator learning with a certain number of examples. Its capability to adapt to varying numbers of examples comes from the inclusion of different numbers of examples in distinct prompt instances.

We identified this architecture and training method as inefficient. In this paper, we propose to merge the encoder and decoder into a single transformer, so that we can train a language-model-like architecture, or directly fine-tune existing language models, for in-context operator
learning. Moreover, we train the model to execute in-context operator learning in an autoregressive manner, i.e., the model predicts the QoI in each example based on previous examples, similar to the ``next-token prediction'' training scheme in language models. Such ``next-function prediction'' training scheme is more efficient, as for each prompt instance, the model simultaneously performs in-context operator learning with varying numbers of examples, ranging from zero (with a caption) or one (without a caption) up to the maximum capacity.

While drawing parallels between language models and operator learning, it's essential to underscore the distinctions and unique requirements of operator learning: (1) The model should be invariant to the permutation of tokens within a function, since these tokens are unordered. (2) The prediction of a QoI function should not be limited to a preset collection of function inputs but is applicable to any inputs. (3) For a QoI function, the outputs corresponding to specific queries should not be generated sequentially as in language models. Rather, these predictions should be made in parallel and independent of each other. 

To address these requirements, we need to design customized input and output sequences, as well as a transformer mask, which will be discussed in the following subsections.

\subsection{Input Tokens}

The input sequence consists of the caption tokens and function tokens constructed from the key-value pairs. Here the key consists of the temporal and spatial coordinates, as well as a ``term'' indicator, as aforementioned.

In ICON each token is constructed by concatenating the key, value, as well as a one-hot index vector. Here, we recognize that the one-hot index vector is equivalent to adding the positional encoding after the linear embedding layer. We thus delete the index vector and only use the key and value in function tokens for simplicity. 

The model predicts the QoI function based on the previous examples. Crucially, these predictions should be made for any function inputs, in parallel and independent of each other. To address such requirements, in addition to the condition function tokens and QoI function tokens, we also include the ``query tokens'' in the sequence, which are the vectors representing the keys of the QoI function. Unlike the approach in the encoder-decoder ICON, where queries are created solely for the last example, in our method, queries are created for each example.

As a demonstration, in Table \ref{tab:sec2_demok_linearode}, we show the tokens of the $j$-th example for the one-dimensional forward ODE problem, where the condition consists of the control $c\colon [0,T]\to \R$ and the initial condition $u(0)$; the QoI is the state $u \colon [0,T]\to \R$.  In the table, each column represents a token. We use $n_j-1$ key-value pairs to represent $c$, one key-value pair for $u(0)$,  and $m_j$ key-value pairs for $u$. In the first row, we use the indicators $0$ and $1$ to distinguish different terms in the condition, i.e., $c$ and $u(0)$. The third row is populated with zeros since there are no spatial coordinates in this problem. During training, the keys in query tokens are the same as those for QoIs, but the values are populated with zero.

\subsection{Model and Transformer Mask}\label{sec:method_mask}

Every condition, QoI, and query token is transformed into an embedding vector via a shared embedding layer (e.g., a linear layer or a shallow multi-layer perceptron). These embedding vectors are then concatenated and appended to the caption embeddings, which are transformed from caption tokens with a different embedding layer. 

Before being supplied to the transformer, the input sequence is added with positional encoding. For caption embeddings, the positional encoding mirrors that utilized in language models. To distinguish different examples and different types of tokens (whether condition, QoI, or query), we also introduce learnable positional encoding for the function embeddings. Notably, to ensure the model remains invariant to the order of key-value pairs within a function, every token of the same type within an example shares the same positional encoding. As an example, for five distinct condition-QoI examples, there would be a total of 15 learnable vectors designated for positional encoding: five each for condition tokens, QoI tokens, and query tokens.

\begin{figure}[ht]
% \vskip -0.1in
\centering
\includegraphics[width=0.9\columnwidth]{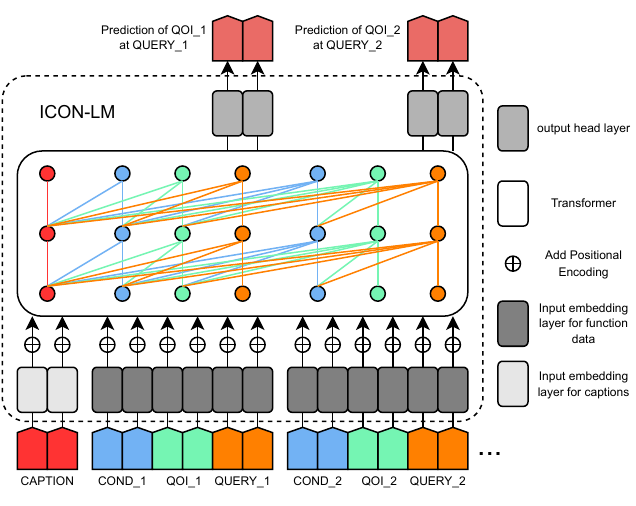}
\vskip -0.1in
\caption{Depiction of the input/output sequence and model architecture of ICON-LM. The connections in the transformer block are a simplified illustration of the attention mask.}
\label{fig:ICON-LM-Model}
% \vskip -0.1in
\end{figure}

After being added with positional encoding, the input sequence is supplied to a transformer. The output sequence of the transformer is then fed into a head layer (e.g., a linear layer or a shallow multi-layer perceptron), to align the dimensions with those of the QoI values. 

In the output sequence, we only keep the ones corresponding to the query tokens, since these parts aim to predict the QoI function values evaluated at the queries. For example, if the QoI represents function $u$, the output corresponding to the query token $x$ aims to predict $u(x)$. 

The input/output sequence and the model architecture are depicted in Figure~\ref{fig:ICON-LM-Model}.

The design of the transformer mask is the key challenge in the ICON-LM model due to the following constraints. (1) The model predicts the QoI value, taking into account all the caption tokens, all the conditions and QoI tokens from previous examples, all the condition tokens of the current example, as well as the current query token. (2) When making the prediction, it's crucial to prevent inadvertent leakage of the QoI tokens in the current example, which contain the prediction targets. (3) Also, the queries should not attend to each other, as the predictions should be independent. (4) The invariance to the permutation of tokens within a function should be maintained. (5) In the end, the mask, denoted by $M$, should satisfy $MM=M$, which ensures that there is no unintentional information leakage due to indirect attention, e.g., the token $c$'s information is leaked to the token $a$ through $a$ attending to $b$, and $b$ attending to $c$.

\begin{figure}[ht]
\centering
\includegraphics[width=\columnwidth]{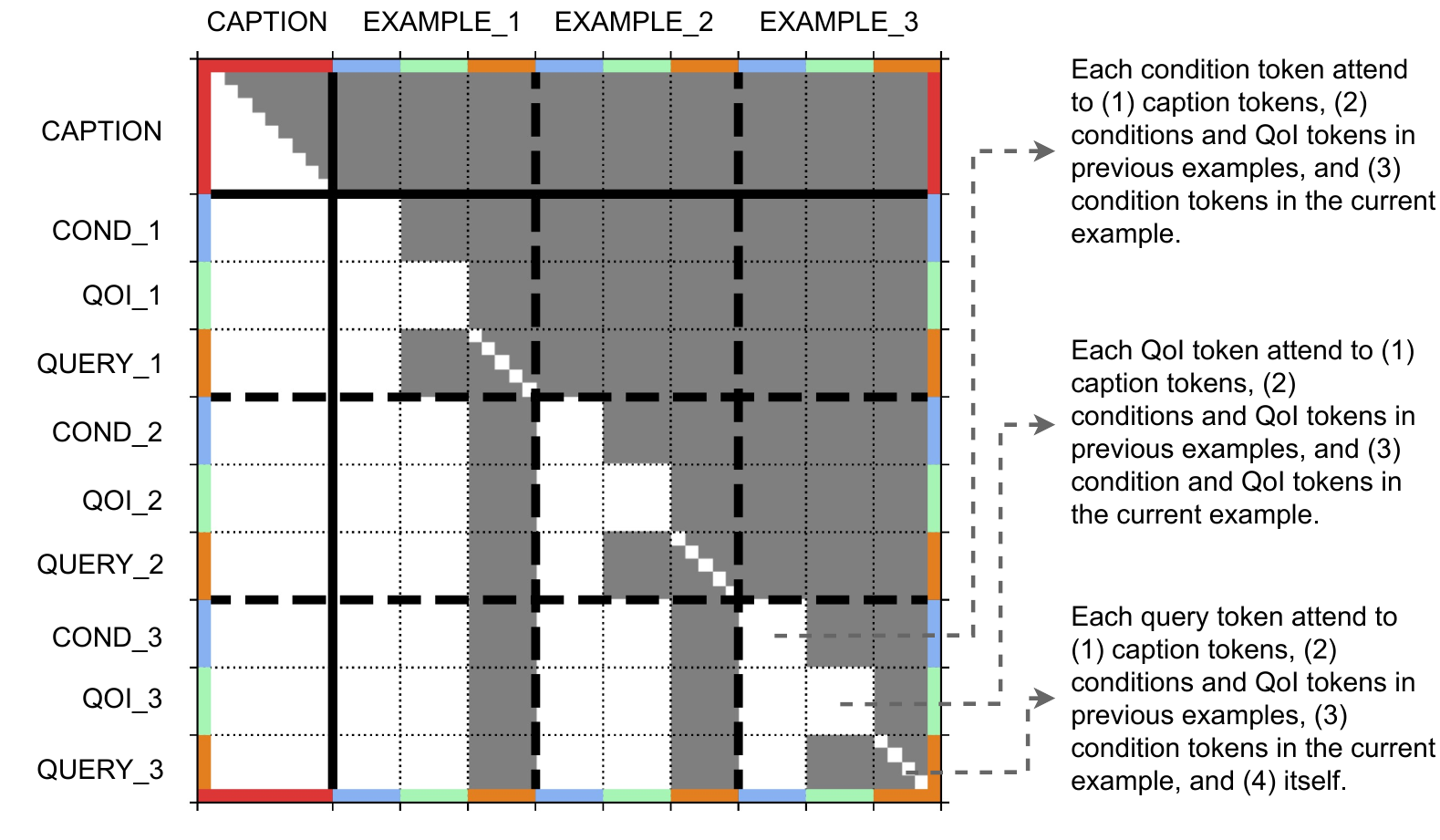} 
\caption{The transformer mask for ICON-LM with three condition-QoI pairs. White cells representing ones, and grey cells representing zeros. 
}
\label{fig:ICON-LM-Mask}
\vskip -0.1in
\end{figure}

We carefully designed the mask that satisfies all the constraints above, illustrated in Figure~\ref{fig:ICON-LM-Mask}. The mask block for caption tokens is lower triangular, in consistency with the existing generative language model. The other blocks are not lower triangular for the sake of permutation invariance. The blocks for queries are diagonal, indicating that the query tokens do not attend to each other.

% Due to various lengths of captions and key-value pairs within a function, we may have to include placeholder tokens in the input sequence to maintain a consistent length in a batch. These placeholder tokens need to be ignored in the attention mechanism. We accomplish this by carrying out an element-wise multiplication of the previously constructed mask with another mask specifically designed to indicate these placeholder tokens, in which the columns corresponding to the placeholder tokens are set to zero, and the rest are set to one.

Since the model architecture is similar to that of language models, we can directly fine-tune existing language models for in-context operator learning, especially for multi-modal learning. The only changes are the embedding layer for function tokens and the head layer, as well as the customized transformer mask.

\subsection{Training and Inference}

We can train the ICON-LM model to execute in-context operator learning, with the option of including or excluding captions. The loss function is the mean squared error between the predicted QoI values and the actual labels. For training inclusive of captions, the loss function is calculated from the first example prediction up to the last, with the first example prediction being a zero-shot -- a prediction solely based on the caption and condition, excluding any other examples. When training without captions, we exclude the caption from the input sequence and calculate the loss function from the second example's predictions to the last, bypassing zero-shot learning as predicting the QoI value without any example or caption is not meaningful. The total loss for multi-modal training comprises the losses from both options.

During inference, provided with a few example condition-QoI pairs and a question condition, we want to predict the QoI corresponding to the question condition. Importantly, the prediction of the QoI should be feasible at any location within the domain, rather than being limited to predetermined fixed positions. ICON-LM efficiently achieves this by constructing question query tokens, where the keys represent where we aim to evaluate the predicted QoI. Owing to a carefully designed mask, these query tokens operate independently, and a flexible number of them is allowed. The question condition and query tokens are then appended to the input sequence as the ``last example''. The QoI tokens for the question are not required, so are all the query tokens in examples since we don't need to predict example QoIs during inference. 

% Since some components of the input sequence can, or should, be excluded in certain scenarios. This affects the corresponding rows and columns of the masks as well.

% In this paper, we applied the following four variations:

% \begin{enumerate}[leftmargin=0\textwidth]
%     \item[] \textbf{Training with caption}: Here, predictions are needed for all queries. While the full mask described in \cref{sec:method_mask} is applicable, we find that the QoI tokens in the last example are never attended by other tokens. Hence, they are omitted from the input sequence to minimize the computational load.
%     \item[] \textbf{Training without caption}: In this scenario, predictions are required for all queries, except those in the first example. We exclude the caption tokens, the query tokens in the first example, and the QoI tokens in the last example from the input sequence.
%     \item[] \textbf{Inference with caption}: Here, the focus lies only on the prediction for ``the last example''. As such, the query tokens from all preceding examples can be omitted from the input sequence, along with the last example's QoI tokens, which are absent during the inference stage.
%     \item[] \textbf{Inference without caption}: This case parallels the previous ``inference with caption'' scenario, except for the omission of the caption tokens from the input sequence.
% \end{enumerate}

% These four mask variations are depicted in Figure~\ref{fig:mask_variants}.

\section{Experiments}\label{sec:exp}

\subsection{ICON-LM v.s. Encoder-Decoder ICON}\label{sec:exp_compare}

This section serves to show the performance of ICON-LM for single-modal in-context operator learning, i.e., without captions, and compare it with the baseline, the encoder-decoder ICON model. 

\begin{figure}[!ht]
\centering
\includegraphics[width=0.8\columnwidth]{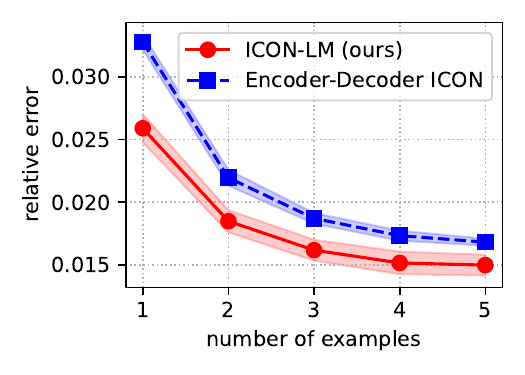}
\vskip -0.1in
\caption{Comparison of ICON-LM (ours) and encoder-decoder ICON for single-modal in-context operator learning. We calculate the relative testing error averaged over all 19 types of problems, and take the mean and standard deviation over three runs, shown as the solid line and the shaded area, respectively.}
\label{fig:compare}
% \vskip -0.1in
\end{figure}

\begin{figure*}[!ht]
\centering
\begin{subfigure}[b]{0.24\textwidth}
\includegraphics[width=\textwidth]{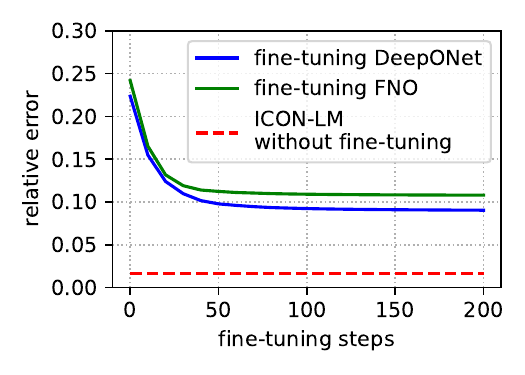}
\vskip -0.1in
\caption{}
\label{fig:compare_deepo_curve}
\end{subfigure}
\begin{subfigure}[b]{0.24\textwidth}
\includegraphics[width=\textwidth]{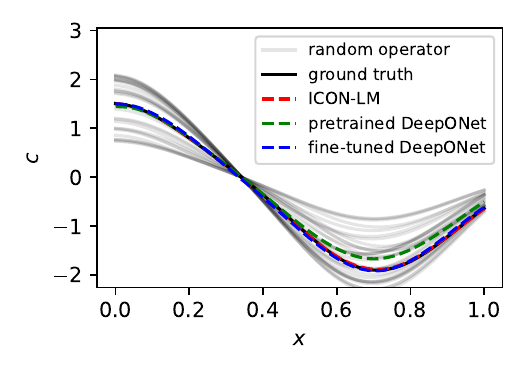}
\vskip -0.1in
\caption{}
\label{fig:compare_deepo_error_small}
\end{subfigure}
\begin{subfigure}[b]{0.24\textwidth}
\includegraphics[width=\textwidth]{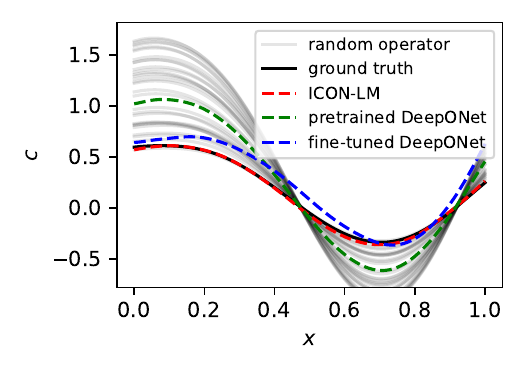}
\vskip -0.1in
\caption{}
\label{fig:compare_deepo_error_large}
\end{subfigure}
\begin{subfigure}[b]{0.24\textwidth}
\includegraphics[width=\textwidth]{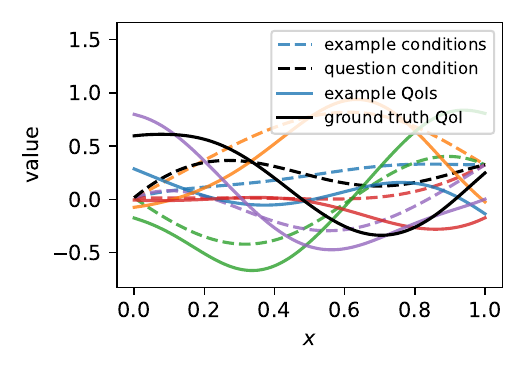}
\vskip -0.1in
\caption{}
\label{fig:compare_deepo_data}
\end{subfigure}
\vskip -0.1in
\caption{Comparison of ICON-LM against FNO and DeepONet. (a) relative error during fine-tuning FNO and DeepNet. (b) prediction for a testing operator close to the mean operator. (c) prediction for a testing operator far from the mean operator. (d) five examples and the question condition for the testing operator in (c).
}
\label{fig:compare_deepo}
\vskip -0.1in
\end{figure*}

% We inherit the original training configurations for the encoder-decoder ICON. Specifically, 
The encoder-decoder ICON is trained with $J$ examples alongside one question, with $J$ randomly selected between one and five for each prompt instance. In contrast, ICON-LM is trained with six examples per instance, allowing for concurrent one-shot to five-shot learning. During both training and testing, each condition or QoI has 41 to 50 tokens. We emphasize that although the number of examples varies between the two models, the total training dataset remains consistent for both, which are presented in \cref{sec:data}.

Both models are trained with the same setups for optimizer and learning rate schedule. More details on the model sizes and training configurations are given in \cref{sec:ap_config}. The encoder-decoder ICON encompasses approximately 31.6 million parameters, whereas the ICON-LM has nearly half that number, at around 15.8 million. This substantial reduction is credited to the ICON-LM's simplified architecture, which employs a single transformer roughly equivalent in size to the encoder or decoder in the baseline ICON. Compared with the baseline ICON, the larger sequence length (about $\times 1.5$) in ICON-LM requires more GPU memory, but this is largely offset by the single transformer design, and slightly smaller batch size (32 for encoder-decoder ICON, and 24 for ICON-LM). With such setups, both models take about 19GB GPU memory, and can fit in one NVIDIA GeForce RTX 4090 GPU with 24 GB memory. As for the time consumption, the training takes about 41.5 hours for the encoder-decoder ICON, and about 37.5 hours for ICON-LM.

We compare the relative testing error averaged over all 19 types of problems from one-shot to five-shot learning. The results are shown in Figure~\ref{fig:compare}. It's clear that ICON-LM consistently outperforms baseline encoder-decoder ICON.

\subsection{In-Context Operator Learning v.s. Classic Operator Learning}

How does in-context operator learning compare with classic operator learning, especially when the data is limited? We examine ICON-LM against classic operator learning methods, including FNO \cite{li2021fourier} and DeepONet \cite{lu2021learning}. Specifically, we use the example of Problem \#14 out of 19 types, namely the inverse nonlinear reaction-diffusion PDE problem, and focus on the five-shot learning scenario. Each condition $u(x)$ and QoI $c(x)$ is represented by 101 evenly spaced data points in $x\in[0,1]$.

The ICON-LM model is inherited from \cref{sec:exp_compare}. While trained with 41 to 50 tokens for each condition/QoI, it can generalized to 101 tokens without any fine-tuning. To enable few-shot learning, we pretrain the FNO and DeepONet models using the training dataset designated for Problem \#14, which comprises a distribution of operators denoted as $\mathcal{P}$. In the pretraining, these models aim to approximate the mean operator and predict $\mathbb{E}_{T\sim \mathcal{P}} T(u)$ for a given condition $u$. Then for each testing operator $T^* \sim \mathcal{P}$, we fine-tune the pretrained models using five examples to approximate $T^*$. The details of operator learning models, pretraining and fine-tuning configurations are in \cref{sec:ap_config}. We note that the parameter numbers of FNO and DeepONet are comparable to or slightly larger than ICON-LM.
The comparison is illustrated in Figure~\ref{fig:compare_deepo}.

The pretrained operator learning models effectively approximate the mean operator, and as a result, the fine-tuned models show a satisfactory approximation of the testing operator $T^*$ when $T^*$ is close to the mean operator, as shown in Figure~\ref{fig:compare_deepo_error_small}. However, their performance deteriorates when attempting to approximate $T^*$ that deviates from the mean, as shown in Figure~\ref{fig:compare_deepo_error_large}. In contrast, ICON-LM consistently performs well due to its in-context learning capability for a distribution of operators. This is highlighted when considering the relative error averaged over 500 testing cases, as depicted in Figure~\ref{fig:compare_deepo_curve}. The error of ICON-LM with just one forward pass for each testing case is substantially lower compared with FNO and DeepONet fine-tuned for each testing case.

\subsection{Multi-Modal In-Context Operator Learning}\label{sec:exp_multi}

In this section, we demonstrate the effectiveness of multi-modal in-context operator learning. We full-parameter fine-tuned the GPT-2 model with 124M parameters~\cite{radford2019language} in the ICON-LM framework. The training setup is the same as ICON-LM in \cref{sec:exp_compare}, except that the total loss for multi-modal training comprises the losses with captions and without captions, and the batch size is 10. The training takes about 5.5 days on dual NVIDIA GeForce RTX 4090 GPUs.

\begin{figure}[!ht]
\centering
\includegraphics[width=0.8\columnwidth]{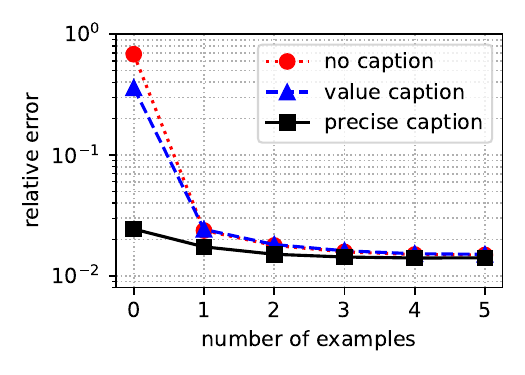}
\vskip -0.2in
\caption{Comparison of multi-modal ICON-LM with vague captions, precise captions, and no captions.}
\label{fig:cap_vs_nocap}
\vskip -0.1in
\end{figure}

In Figure~\ref{fig:cap_vs_nocap}, we show the relative testing errors averaged over all 19 types of problems from one-shot to five-shot learning, examining the performance when prompted without captions, with vague captions, and with precise captions. A comprehensive comparison for each specific problem type can be found in the \cref{sec:ap_details}.

It's clear that the testing error decreases as we move from no caption to vague caption to precise caption, especially for zero-shot. It is worth noting that zero-shot learning with vague captions performs surprisingly well for some problems. 
% These include Problem \#7, \#8, \#10, and \#17.
% These include the forward and inverse damped oscillator series problems (\#7,\#8), the inverse Poisson equation problem (\#10), and the mean-field control problem with $g$ parameters, mapping from 2D density field to 2D density field (\#17). 
These results are not as surprising upon closer examination, since the operator parameters for these problems can be inferred from the conditions provided, once the problem type is identified with the vague captions. For instance, the decay rate of the damped oscillator can be computed from a segment of the time series, the boundary conditions of the Poisson equation are encapsulated within the conditions of $u(x)$, and the terminal cost in the mean-field control problem can be derived from the density field during the interval $t\in[0,0.5]$. These examples highlight the impressive capabilities of the ICON-LM framework in combining multi-modal information to learn the operator.

\subsection{Ablation Study}\label{sec:exp_ablation}

Here we conduct an ablation study to inspect the differences between pretrained and unpretrained language models within the multi-modal in-context operator learning framework. The training setup is the same as in \cref{sec:exp_multi},  with the sole variable being the initial state of the GPT-2 model: either pretrained or initialized randomly. 

In Table~\ref{tab:pretrained_vs_unpretrained} we show the results for zero-shot learning with precise captions, where the relative error is averaged across all 19 problem types. The results reveal that while the unpretrained GPT-2 model performs better on the training captions, it generalizes poorly to the testing captions. Meanwhile, the generalization gap is much smaller for the pretrained GPT-2 model. This shows that fine-tuning a pretrained language model enhances the ICON-LM's proficiency in executing multi-modal tasks related to scientific machine learning.

% \vskip -0.15in
\begin{table}[!ht]
\caption{Zero-shot relative errors of ICON-LM with pretrained and unpretrained GPT-2 model.}
\label{tab:pretrained_vs_unpretrained}
\vskip 0.15in
\begin{center}
\begin{small}
\begin{sc}
\begin{tabular}{lccc}
\toprule
Model & Train (\%) & Test (\%)& Gap (\%) \\
\midrule
Pretrained  & $2.30$ & \textbf{2.44} & \textbf{0.14} \\
Unpretrained & \textbf{2.07}& $3.08$& $1.01$\\
\bottomrule
\end{tabular}
\end{sc}
\end{small}
\end{center}
\vskip -0.1in
\end{table}
\section{Summary}\label{sec:sum}

We present a novel approach to transform in-context operator learning for scientific machine learning into a multi-modal framework, meaning that the model can learn the operator from captions, function data, or a combination of both. These captions incorporate human knowledge about the operator, in the form of natural language descriptions and equations. We also introduce a more efficient model architecture for multi-modal in-context operator learning, namely ``ICON-LM''. This architecture closely aligns with language models, with carefully designed input sequences and transformer masks.

In the experiments, we compared the ICON-LM model with the baseline encoder-decoder ICON model, in the single-modal learning scenario. The proposed ICON-LM model, comprising approximately half parameters, surpasses the performance of the baseline ICON model with less training time. We also showed the advantage of ICON-LM for in-context operator learning compared with FNO and DeepONet for classic operator learning, especially when the data is limited for testing operators.
% This can be attributed that the encoder-decoder ICON model is trained with a fixed number of examples in each step, while the ICON-LM model is trained with varying numbers of examples concurrently in each training step.

We also fine-tuned GPT-2 model for multi-modal in-context operator learning. We found that captions' presence, especially precise ones that disclose the parameters in the operators, significantly improved learning performance when the number of examples was limited. The model performance is improved even with vague captions that only disclose the operator type, showing impressive capabilities of the ICON-LM framework in combining multi-modal information to learn the operator. 

In this paper, we focused on the transition to the multi-modal framework and the autoregressive training scheme, with the function representation limited to scattered points. The proposed multi-modal autoregressive training scheme should be able to generalize to other function representations, which are left to future work.

\section*{Impact Statements}
This paper presents work whose goal is to advance the field of Machine Learning. There are many potential societal consequences of our work, none which we feel must be specifically highlighted here.


\bibliography{biblist}
\bibliographystyle{icml2024}

\newpage
\appendix
\onecolumn

\section{Problems}\label{sec:ap_problem}
\begin{figure}[!ht]
\centering
\includegraphics[width=0.17\textwidth]{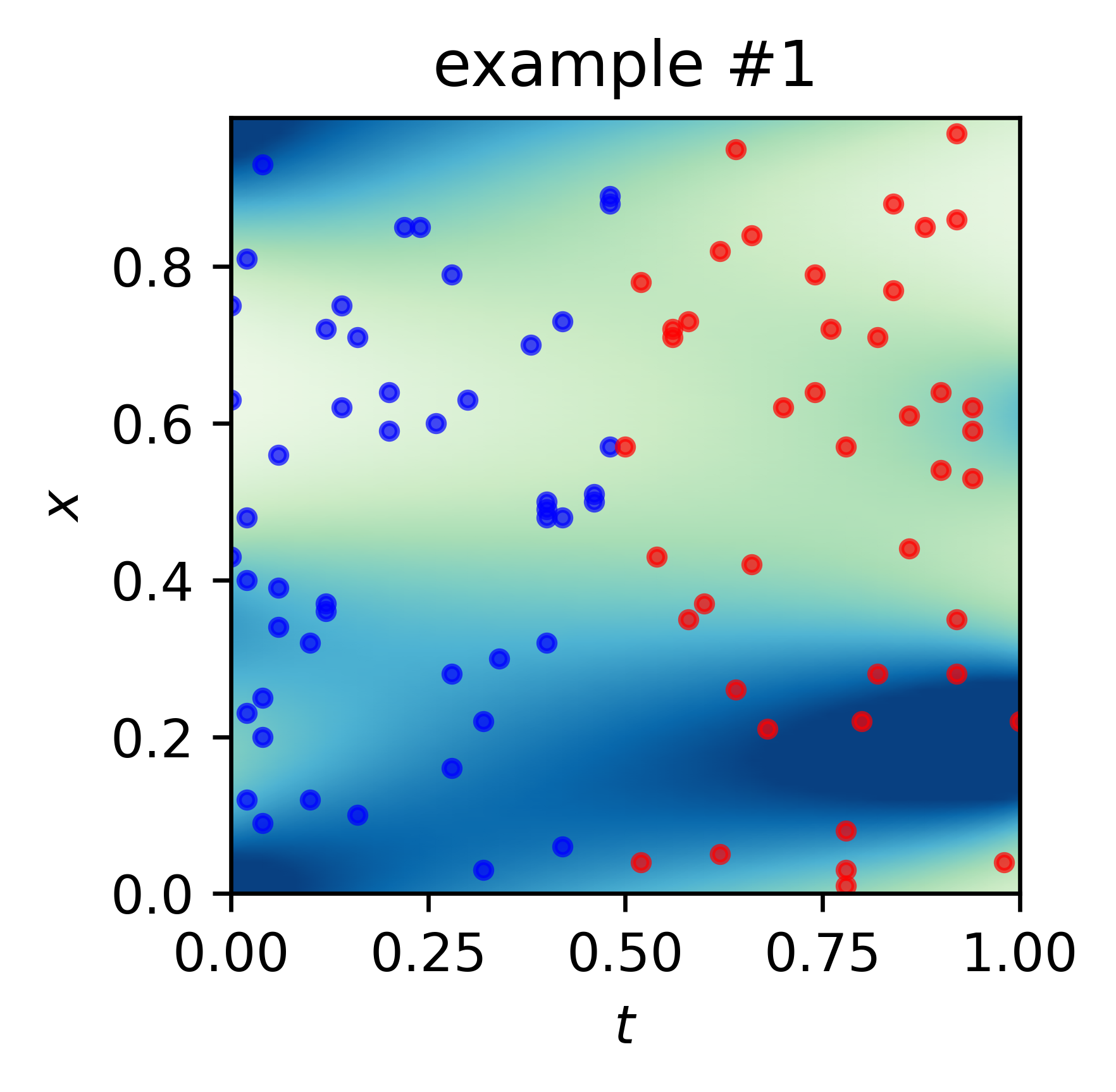}
\includegraphics[width=0.17\textwidth]{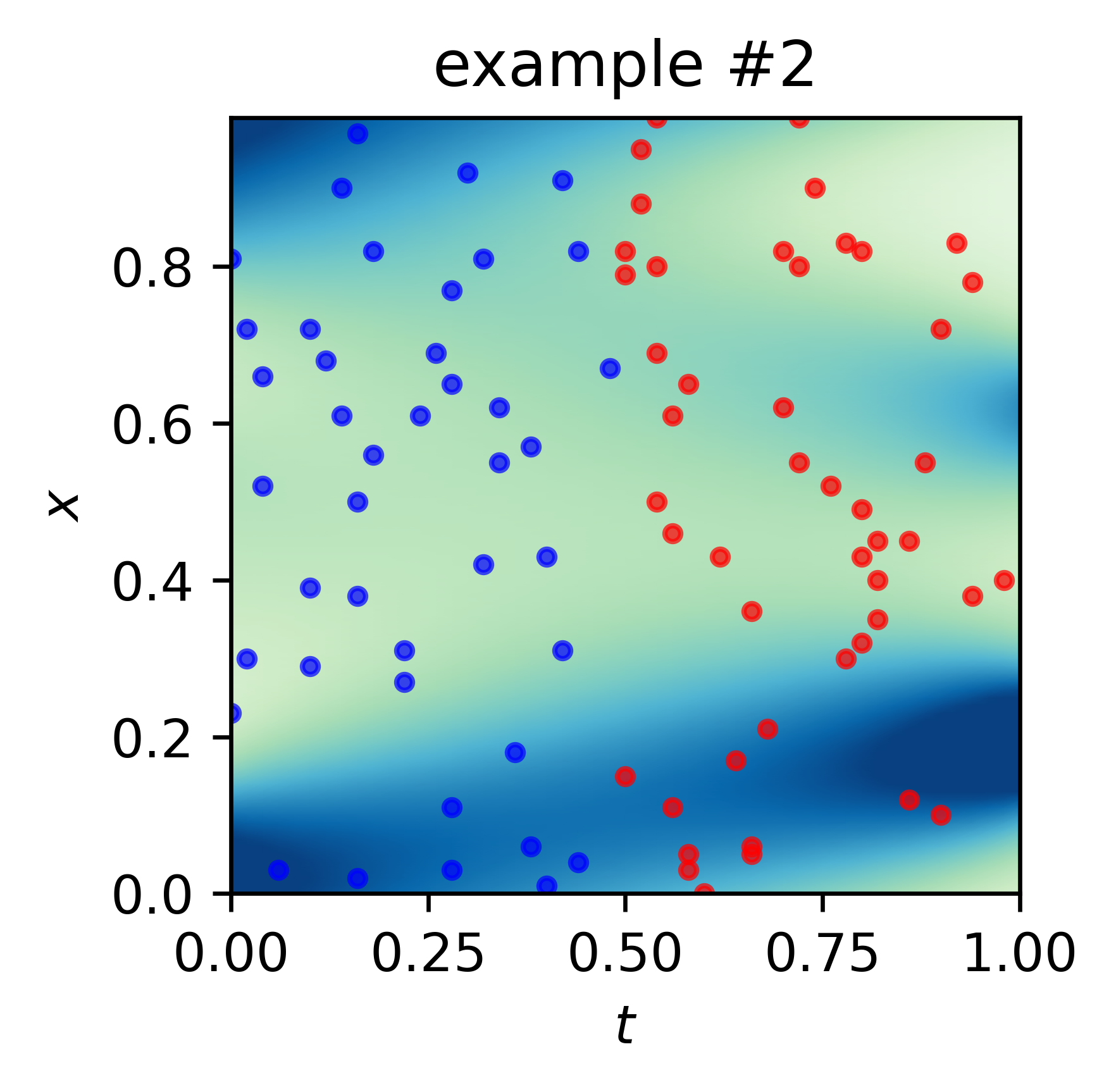}
\includegraphics[width=0.17\textwidth]{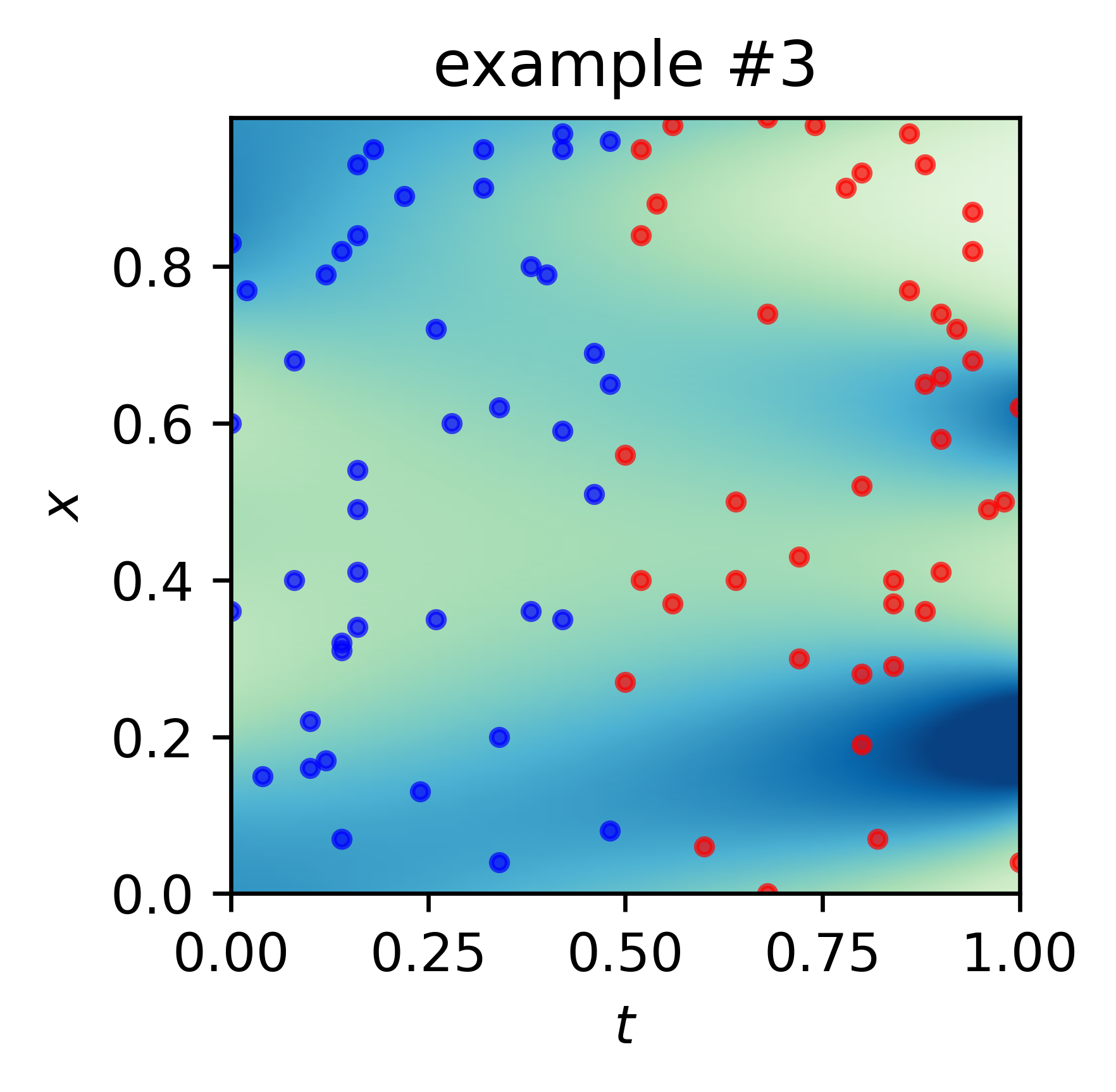}
\includegraphics[width=0.45\textwidth]{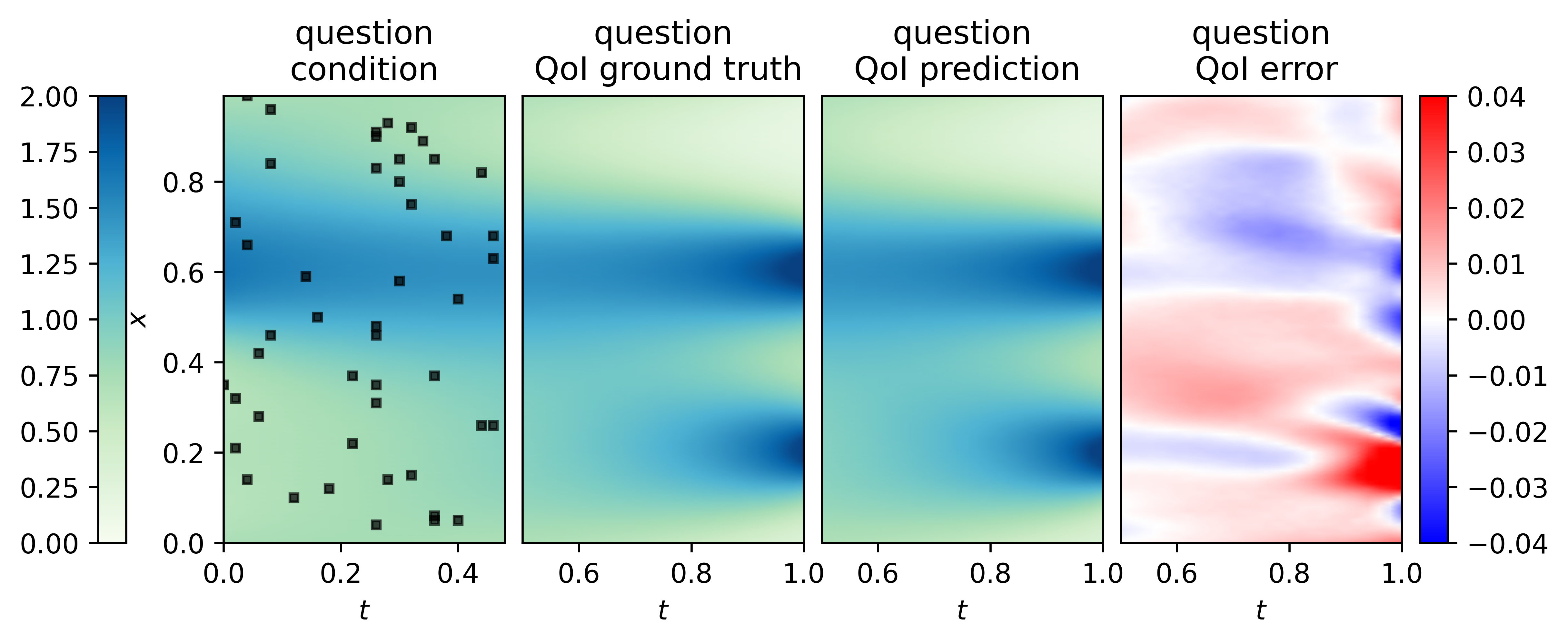}
\vskip -0.1in
\caption{Illustration of an in-context learning case in Problem \#17. The blue/red round dots represent the data points for example conditions/QoIs in the prompt; the black square dots represent the data points for the question condition in the prompt.}
\label{fig:2d_profile}
\vskip -0.1in
\end{figure}

Here we list the 19 types of differential equation problems in the dataset. An instance of Problem \#17 is illustrated in Figure~\ref{fig:2d_profile}. In this case, we need to infer the operator defined by the hidden terminal cost $g(x)$ from merely three examples (without captions), and then apply it to the question condition, the density field during $t\in[0,0.5]$, to make predictions of the density field during a later time interval. Note that in the prompt, there are only 41 to 50 scattered data points for each condition/QoI function, but we can make the prediction in the whole domain of $(t,x)\in[0.5,1]\times[0,1]$. 

% For more details of these problems and the dataset, readers are referred to the main text and supporting information of \cite{yang2023context}.

\begin{table}[!htbp]
\scriptsize
\centering
\caption{List of differential equation problems.}
\vspace{0.15in}
\renewcommand{\arraystretch}{1.5}
\begin{tabular}{|p{0.15cm}|p{3.65cm}|p{4.7cm}|p{1.6cm}|p{2.3cm}|p{2.3cm}|}
% \begin{tabular}{|l|l|l|l|l|l|}
\hline
\# & Problem Description & \text{Differential Equations} & Parameters & Conditions & QoIs \\
\hline
1 & Forward problem of ODE 1 & \multirow{2}{*}{{\parbox{6cm}{$\frac{d}{dt}u(t) = a_1 c(t) + a_2$\\ for $t\in[0,1]$}}} & \multirow{2}{*}{$a_1, a_2$} & $u(0), c(t), t\in[0,1]$ & $u(t), t\in[0,1]$ \\
\cline{1-2}\cline{5-6}
2 & Inverse problem of ODE 1 & & & $u(t), t\in[0,1]$ & $c(t), t\in[0,1]$ \\
\hline
3 & Forward problem of ODE 2 & \multirow{2}{*}{{\parbox{6cm}{$\frac{d}{dt}u(t) = a_1 c(t) u(t) + a_2$\\ for $t\in[0,1]$}}} & \multirow{2}{*}{$a_1, a_2$} & $u(0),c(t), t\in[0,1]$ & $u(t), t\in[0,1]$ \\
\cline{1-2}\cline{5-6}
4 & Inverse problem of ODE 2 & & & $u(t), t\in[0,1]$ & $c(t), t\in[0,1]$ \\
\hline
5 & Forward problem of ODE 3 & \multirow{2}{*}{{\parbox{6cm}{$\frac{d}{dt}u(t) = a_1 u(t) + a_2 c(t) + a_3$\\ for $t\in[0,1]$}}} & \multirow{2}{*}{$a_1, a_2, a_3$} & $u(0),c(t), t\in[0,1]$ & $u(t), t\in[0,1]$ \\
\cline{1-2}\cline{5-6}
6 & Inverse problem of ODE 3 & & & $u(t), t\in[0,1]$ & $c(t), t\in[0,1]$ \\
\hline
7 & Forward damped oscillator  & \multirow{2}{*}{{\parbox{6cm}{$u(t) = A \sin(\frac{2\pi}{T} t + \eta)e^{-kt}$\\ for $t\in[0,1]$}}} & \multirow{2}{*}{$k$} & $u(t), t \in[0,0.5)$ & $u(t), t \in[0.5 , 1]$ \\
\cline{1-2}\cline{5-6}
8 & Inverse damped oscillator  & & & $u(t), t\in[0.5 , 1]$ & $u(t), t\in[0 , 0.5)$ \\
\hline
9 & Forward Poisson equation & \multirow{2}{*}{$\frac{d^2}{dx^2}u(x) = c(x)$ for $x\in[0,1]$} & \multirow{2}{*}{$u(0), u(1)$} & $c(x), x\in[0,1]$ & $u(x), x\in[0,1]$ \\
\cline{1-2}\cline{5-6}
10 & Inverse Poisson equation & & & $u(x), x\in[0,1]$ & $c(x), x\in[0,1]$ \\
\hline
11 & Forward linear reaction-diffusion & \multirow{2}{*}{{\parbox{6cm}{$-\lambda a\frac{d^2}{dx^2}u(x) + k(x)u(x) = c$ \\ for $x\in[0,1]$, $\lambda = 0.05$}}} & \multirow{2}{*}{$u(0), u(1), a,c$} & $k(x), x\in[0,1]$ & $u(x), x\in[0,1]$ \\
\cline{1-2}\cline{5-6}
12 & Inverse linear reaction-diffusion & & & $u(x), x\in[0,1]$ & $k(x), x\in[0,1]$ \\
\hline
13 & Forward nonlinear reaction-diffusion  & \multirow{2}{*}{{\parbox{6cm}{$- \lambda a \frac{d^2}{dx^2}u(x) + ku(x)^3 = c(x)$ \\for $x\in[0,1]$, $\lambda = 0.1$}}} & \multirow{2}{*}{$u(0), u(1), k, a$} & $c(x), x\in[0,1]$ & $u(x), x\in[0,1]$ \\
\cline{1-2}\cline{5-6}
14 & Inverse nonlinear reaction-diffusion & & & $u(x), x\in[0,1]$ & $c(x), x\in[0,1]$ \\
\hline
 15 &   MFC $g$-parameter $1$D $\rightarrow 1$D & \multirow{5}{*}{\parbox{6cm}
{$\quad$\\
$\inf_{\substack{\rho, m}}\iint c\dfrac{m^2}{2 \rho} dx dt + \int g(x) \rho(1,x) dx$\\
s.t.
$\partial_t \rho(t,x) + \nabla_x\cdot m(t,x) = \mu \Delta_x \rho(t,x)$
\\
for $t \in [0,1], x \in [0,1]$,\\
$c = 20, \mu = 0.02$, \\
periodic spatial boundary condition
 }} & \multirow{5}{*}{$g(x), x\in[0,1]$} & {\parbox{6cm}{ $\rho(t=0,x)$, $x\in[0,1]$}}& {\parbox{6cm}{$\rho(t=1,x), x\in[0,1]$}} \\
    \cline{1-2}\cline{5-6}
16 &     MFC $g$-parameter $1$D $\rightarrow 2$D  &  &  & {\parbox{6cm}{ $\rho(t=0,x)$, $x\in[0,1]$}}&{\parbox{6cm}{$\rho(t,x)$, \\$t\in[0.5,1]$, $x\in[0,1]$}} \\
    \cline{1-2}\cline{5-6}
 17 &    MFC $g$-parameter $2$D $\rightarrow 2$D  &  &  & {\parbox{2.8cm}{$\rho(t,x)$,\\ $t\in[0,0.5)$, $x\in [0,1]$}} & \parbox{2.7cm}{$\rho(t,x)$,\\ $t\in[0.5,1]$, $x\in[0,1]$} \\
    \cline{1-2}\cline{4-6}
18 &   MFC $\rho_0$-parameter $1$D $\rightarrow 1$D  &  & \multirow{2}{*}{{\parbox{2.7cm}{$\rho(t=0,x)$\\ $x\in[0,1]$}}} & \multirow{2}{*}{$g(x), x\in[0,1]$} & {\parbox{6cm}{$\rho(t=1,x)$, $x\in[0,1]$} }\\
    \cline{1-2}\cline{6-6}
 19 &  MFC $\rho_0$-parameter $1$D $\rightarrow 2$D  &  &  &  & {\parbox{2.7cm}{$\rho(t,x)$,\\$t\in[0.5,1]$, $x\in[0,1]$}}\\
\hline
\end{tabular}
\renewcommand{\arraystretch}{1}
\label{tab:problem_table}
\end{table}

\section{Caption Data Generation}\label{sec:ap_caption_gen}

Here's how we generated these captions with the assistance of GPT-4.

First, we utilized the GPT-4 API to create two categories of captions for every problem type. To reduce the workload, we merged the forward and inverse problems for ODEs/PDEs when generating captions, and then manually specified the condition and QoI terms to distinguish between the two. Similarly, for the mean-field control problems we merged the problems with condition/QoI terms in different time intervals. From both the vague and precise categories, we generated 100 captions each, allocating 80 for training and 20 for testing. Most of the AI-produced captions were suitable, while a few necessitated slight manual adjustments.

For the vague group, we instructed GPT-4 to use natural language to describe the equation or tell the form of the equation without revealing the specific parameter values. For the precise group, we instructed GPT-4 to leave placeholders for the actual parameters. 

During training, for each prompt instance, a caption is randomly selected from all the training captions, including vague ones and precise ones, associated with the operator. If a precise caption is selected, the placeholders for parameters are replaced with the actual parameter values. It's worth noting that for mean-field control problems, since the parameters are functional, we adopted a discretization approach to represent them effectively. Similarly, a caption is randomly selected from the testing group for each prompt instance during testing.

\section{Caption Examples}\label{sec:ap_caption_ex}

Here we show several examples of training and testing captions for three characteristic problem types: ODE 3 forward problem, PDE 3 forward problem, and MFC $g$-parameter $1$D $\rightarrow 1$D. For each type, we show four vague captions and four precise captions, with the former two used in training, and the latter two for testing.

\lstset{
    basicstyle=\small\ttfamily,
    breaklines=true,
    % breakindent=0pt,
    columns=flexible,
    keepspaces=true,
    escapeinside={(*@}{@*)}
}

\begin{enumerate}
\item Caption examples for ODE 3 forward problem.
\begin{lstlisting}
----Vague----
Variable $u$'s time derivative is $du(t)/dt = a_1 \cdot u(t) + a_2 \cdot c(t) + a_3$. Condition: $u(0)$ and $c(t), t\in[0,1]$, QoI: $u(t), t\in[0,1]$.
The ordinary differential equation represents the growth rate of variable $u(t)$ in relation to itself and the control function $c(t)$. Condition: $u(0)$ and $c(t), t\in[0,1]$, QoI: $u(t), t\in[0,1]$.
Derivation of $u(t)$ in time following the formula $du/dt = a_1u(t) + a_2c(t) + a_3$. Condition: $u(0)$ and $c(t), t\in[0,1]$, QoI: $u(t), t\in[0,1]$.
An ordinary differential equation with respect to time using a state variable $u(t)$ and a control variable $c(t)$. Condition: $u(0)$ and $c(t), t\in[0,1]$, QoI: $u(t), t\in[0,1]$.

----Precise----
Knowing that $a_1 = -0.0124, a_2 = 1.06, a_3 = 0.105$, the derivative $du(t)/dt = -0.0124 \cdot u(t) + 1.06 * c(t) + 0.105$. Condition: $u(0)$ and $c(t), t\in[0,1]$, QoI: $u(t), t\in[0,1]$.
The state variable changes according to $du(t)/dt = 0.347 \cdot u(t) + 0.535 \cdot c(t) + 0.459$. Condition: $u(0)$ and $c(t), t\in[0,1]$, QoI: $u(t), t\in[0,1]$.
Express an ODE as $\frac{du(t)}{dt} = -0.85 \cdot u(t) + 1.13 \cdot c(t) + -0.779$.  Condition: $u(0)$ and $c(t), t\in[0,1]$, QoI: $u(t), t\in[0,1]$.
This differential equation $du(t)/dt = 0.167 * u(t) + 1.02 * c(t) + 0.457$ shows how $u(t)$ changes with time. Condition: $u(0)$ and $c(t), t\in[0,1]$, QoI: $u(t), t\in[0,1]$.
\end{lstlisting}

\item  Caption examples for PDE 3 forward problem.

\begin{lstlisting}
----Vague----
The nonlinear PDE, written as $-\lambda\frac{d^2u}{dx^2} + a * u^3 = c(x)$, includes the variables $u(x)$ and $c(x)$. Condition: $c(x), x\in[0,1]$, QoI: $u(x), x\in[0,1]$.
The nonlinear PDE, $u''(x) - a \cdot u(x)^3 = c(x)$ roping in $u(x)$ and $c(x)$.  Condition: $c(x), x\in[0,1]$, QoI: $u(x), x\in[0,1]$.
Ponder upon this nonlinear PDE, involving the dependent variable $u(x)$ and the term $c(x)$ constituting the source. Condition: $c(x), x\in[0,1]$, QoI: $u(x), x\in[0,1]$.
This PDE, $-\lambda d^2u/dx^2 + a \cdot u^3 = c(x)$, involves the variables $u(x)$ and $c(x)$. Condition: $c(x), x\in[0,1]$, QoI: $u(x), x\in[0,1]$.

----Precise----
For the given equation $- 0.101 * \frac{d^2u}{dx^2} + 1.16 * u^3 = c(x)$, we have the boundary conditions $u(0) = -0.517$ and $u(1) = -0.689$. Condition: $c(x), x\in[0,1]$, QoI: $u(x), x\in[0,1]$.
The nonlinear PDE is $-0.0504 d^2u/dx^2 + 0.705 \cdot u^3 = c(x)$, with $u(0) = -0.319$ and $u(1) = -0.667$. Condition: $c(x), x\in[0,1]$, QoI: $u(x), x\in[0,1]$.
Equation $- 0.116 \frac{d^2u}{dx^2} + 0.586 \cdot u^3 = c(x)$, is our PDE with $u(0) = 0.322$ and $u(1) = -0.749$. Condition: $c(x), x\in[0,1]$, QoI: $u(x), x\in[0,1]$.
Let's examine this PDE $- 0.139 \frac{d^2u}{dx^2} + 1.25 \cdot u^3 = c(x)$, imposing $u(0) = -0.351$, $u(1) = 0.597$. Condition: $c(x), x\in[0,1]$, QoI: $u(x), x\in[0,1]$.
\end{lstlisting}

\item Caption examples for MFC $g$-parameter $1$D $\rightarrow 1$D.
\begin{lstlisting}
----Vague----
Investigating Mean Field Control Problem involving an interplay between density $\rho$ and an uncertain function $g$ inside the terminal cost. Condition: $\rho(0,x), x\in[0,1]$, QoI: $\rho(1,x), x\in[0,1]$.
In the mean field control problem, we minimize $\int \int \frac{10m^2}{\rho} dx dt + \int g(x)\rho(1,x) dx$, subject to $\partial_{t}\rho + \nabla_{x}\cdot m = 0.02 \Delta_{x}\rho$, with $\rho(0,x)=\rho_{0}(x)$, where $g$ is an unknown function. Condition: $\rho(0,x), x\in[0,1]$, QoI: $\rho(1,x), x\in[0,1]$.
Consider the mean field control problem with density function $\rho(t,x)$ and terminal cost $\int g(x)\rho(1,x) dx$ where $g$ is an unknown function. Condition: $\rho(0,x), x\in[0,1]$, QoI: $\rho(1,x), x\in[0,1]$.
The mean field control problem formulates $\inf_{\rho, m}\iint \frac{10m^2}{\rho} dx dt + \int g(x)\rho(1,x) dx$ subject to $\partial_t \rho(t,x) + \nabla_x m(t,x) = 0.02 \Delta_x \rho(t,x)$, where $g(x)$ is undefined. Condition: $\rho(0,x), x\in[0,1]$, QoI: $\rho(1,x), x\in[0,1]$.

----Precise----
The analysis of $\inf_{\rho, m}\iint \frac{10m^2}{\rho} dx dt + \int g(x)\rho(1,x) dx$ for $t \in [0,1]$, $x \in [0,1]$ and periodic spatial boundary condition, under constraint of $\partial_t \rho(t,x) + \nabla_x m(t,x) = 0.02 \Delta_x \rho(t,x)$, with the function $g$ acting as terminal cost is defined as $g(0), g(0.1), ..., g(0.9)$ = 0.903, 0.957, 0.459, -0.178, -0.83, -1.5, -1.39, -0.189, 0.857, 0.909.  Condition: $\rho(0,x), x\in[0,1]$, QoI: $\rho(1,x), x\in[0,1]$.
Analyzing mean field control problem $\inf_{\rho, m}\iint \frac{10m^2}{\rho} dx dt + \int g(x)\rho(1,x) dx$, subject to the constraints, for $t \in [0,1], x \in [0,1], and terminal function $g$ defined as $g(0), g(0.1), ..., g(0.9) = -0.244, 0.326, 0.598, 0.571, 0.287, 0.0734, 0.00921, -0.299, -0.67, -0.652$. Condition: $\rho(0,x), x\in[0,1]$, QoI: $\rho(1,x), x\in[0,1]$.
We solve a mean field control problem that seeks to minimize $\inf_{\rho, m}\iint \frac{10m^2}{\rho} dx dt + \int g(x)\rho(1,x) dx$ while adhering to $\partial_t \rho(t,x) + \nabla_x m(t,x) = 0.02 \Delta_x \rho(t,x)$ and $\rho(0,x)=\rho_0(x)$. A known function $g$ is given by $g(0), g(0.1), ..., g(0.9)$ = 0.535, 0.976, 1.35, 1.49, 0.135, -1.86, -1.66, -0.692, -0.305, 0.0242. Condition: $\rho(0,x), x\in[0,1]$, QoI: $\rho(1,x), x\in[0,1]$.
Studying the mean field control problem $\inf_{\rho, m}\iint \frac{10m^2}{\rho} dx dt + \int g(x)\rho(1,x) dx$, where $t \in [0,1], x \in [0,1], and $g$ is given as $g(0), g(0.1), ..., g(0.9) = -0.0268, 0.196, 0.08, -0.0463, 0.145, 0.126, 0.169, 0.0845, -0.313, -0.413$. Condition: $\rho(0,x), x\in[0,1]$, QoI: $\rho(1,x), x\in[0,1]$.

\end{lstlisting}

\end{enumerate}

\pagebreak
\section{Neural Network and Training Configurations}\label{sec:ap_config}

The transformer used in \cref{sec:exp_compare} is configured as in Table~\ref{tab:transformer-config}. Both the embedding layer and head layer are linear layers. For fine-tuning the GPT-2 model in \cref{sec:exp_multi}, we apply shallow multilayer perceptrons as the input embedding layer for function data as well as the output head layer, with one hidden layer of dimension 1024. We utilize the AdamW optimizer with a warmup-cosine-decay schedule, employing the configuration in Table~\ref{tab:optimizer-config}.

In \cref{sec:exp_compare}, we pretrained and fine-tuned the FNO and DeepONet. The FNO model is adopted from the official implementation (\url{https://github.com/neuraloperator/neuraloperator}), with \texttt{n\_modes} = 16, \texttt{hidden\_channels} = 512, \texttt{in\_channels} = 2 (one for $x$, one for $u(x)$), and \texttt{out\_channels} = 1. Other parameters are default. The total number of trainable parameters is about 20.2 million. In the DeepONet, both the trunk net and branch net are 6-layer multilayer perceptrons, with hidden and output widths of 1024. The total number of trainable parameters is about 14.8 million. The pretraining of FNO and DeepONet has a batch size of 32, following the same configuration as in Table~\ref{tab:optimizer-config}, except that the total training steps are reduced to 0.1 million, which is sufficient for convergence. As for fine-tuning FNO and DeepONet, we use all the available five examples in the training batch, employing the AdamW optimizer with a constant learning rate 1e-5, weight decay 1e-4, and global norm clip 1.0.

\begin{table}[ht]
\centering
\caption{Configuration of the Transformer in Single-Modal ICON-LM}
\vspace{0.15in}
\begin{tabular}{|c|c|}
\hline
{Layers} & 6 \\
\hline
{Heads in Multi-Head Attention} & 8 \\
\hline
{Input/Output Dimension of Each Layer} & 256 \\
\hline
{Dimension of Query/Key/Value in Attention Function} & 256 \\
\hline
{Hidden Dimension of Feedforward Networks} & 1024 \\
\hline
{Total Trainable Parameters} & 15.8 Million \\
\hline
\end{tabular}
\label{tab:transformer-config}
\end{table}

\begin{table}[ht]
\centering
\caption{Configuration of Optimizer and Learning Rate Schedule}
\vspace{0.15in}
\begin{tabular}{|c|c|}
\hline
{Initial Learning Rate} & 0.0 \\
\hline
{Peak Learning Rate} & 1e-4 \\
\hline
{End Learning Rate} & 0.0 \\
\hline
{Total Training Steps} & 1 million\\
\hline
{Warmup Steps} & First 10\% of Total Steps\\
\hline
{Cosine Annealing Steps} & Remaining Steps\\
\hline
{Global Norm Clip} & 1.0\\
\hline
{Adam $\beta_1$} & 0.9\\
\hline
{Adam $\beta_2$} & 0.999\\
\hline
{Adam Weight Decay} & 1e-4\\
\hline
\end{tabular}
\label{tab:optimizer-config}
\end{table}

% \pagebreak

\section{Detailed Results of Multi-Modal In-Context Operator Learning}\label{sec:ap_details}
\begin{figure}
    \centering
    \begin{subfigure}{.3\textwidth}
        \centering
        \includegraphics[width=\linewidth]{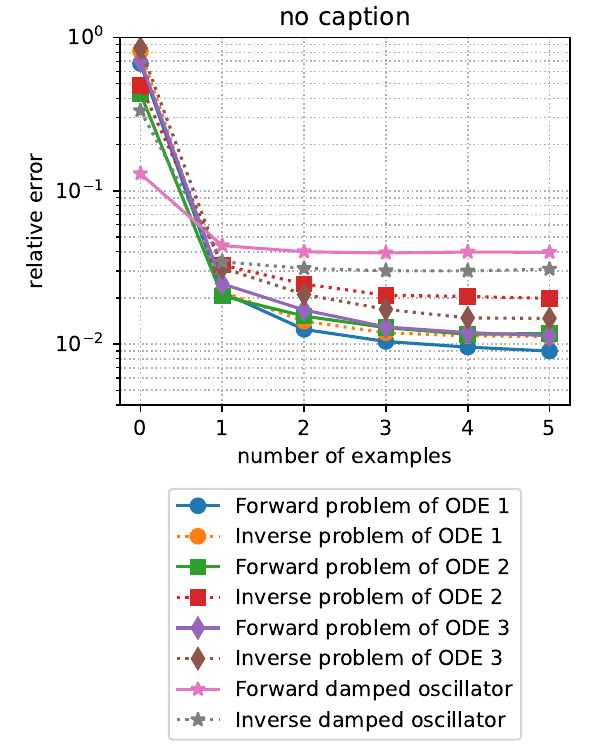}
        \includegraphics[width=\linewidth]{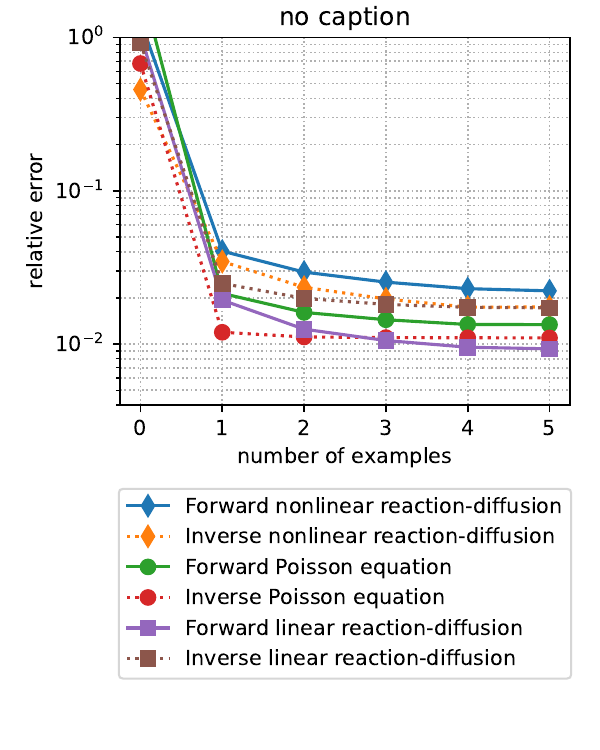}
        \vskip -0.25in
        \includegraphics[width=\linewidth]{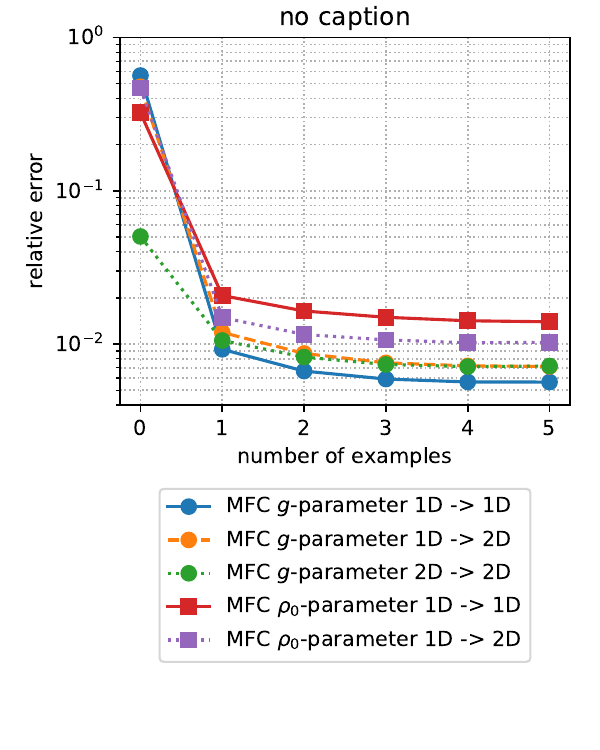}
        \vskip -0.3in
        \caption{}
        \label{fig:caption_error_sub1}
    \end{subfigure}%
    \hfill
    \begin{subfigure}{.3\textwidth}
        \centering
        \includegraphics[width=\linewidth]{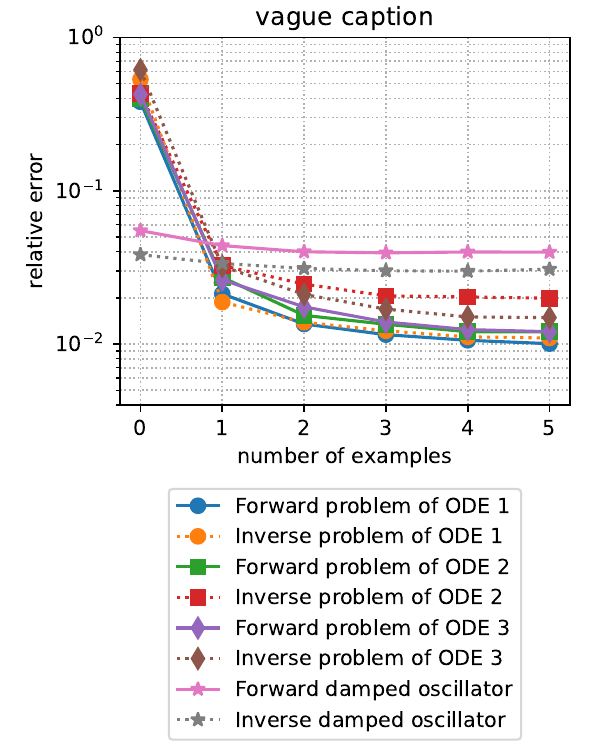}
        \includegraphics[width=\linewidth]{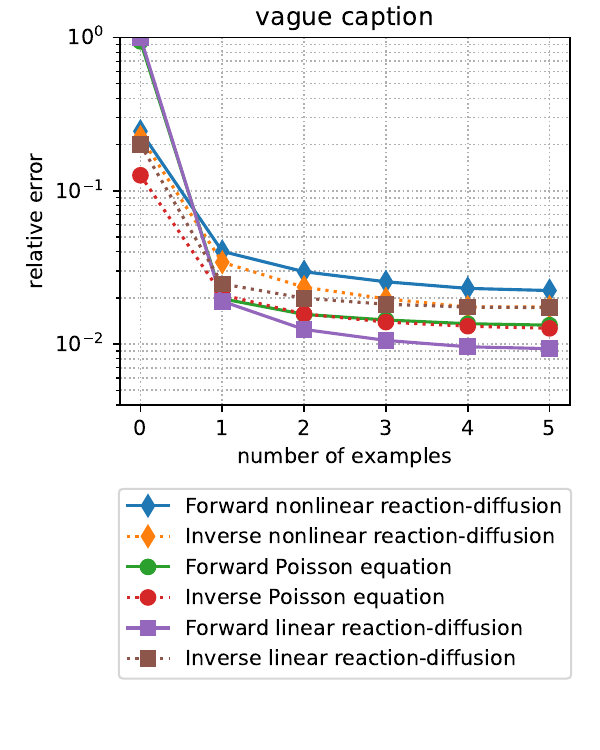}
        \vskip -0.25in
        \includegraphics[width=\linewidth]{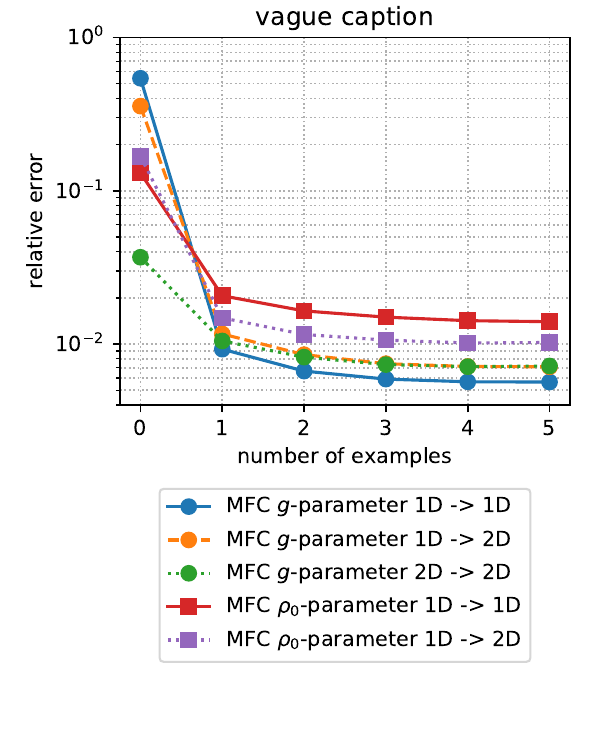}
        \vskip -0.3in
        \caption{}
        \label{fig:caption_error_sub2}
    \end{subfigure}%
    \hfill
    \begin{subfigure}{.3\textwidth}
        \centering
        \includegraphics[width=\linewidth]{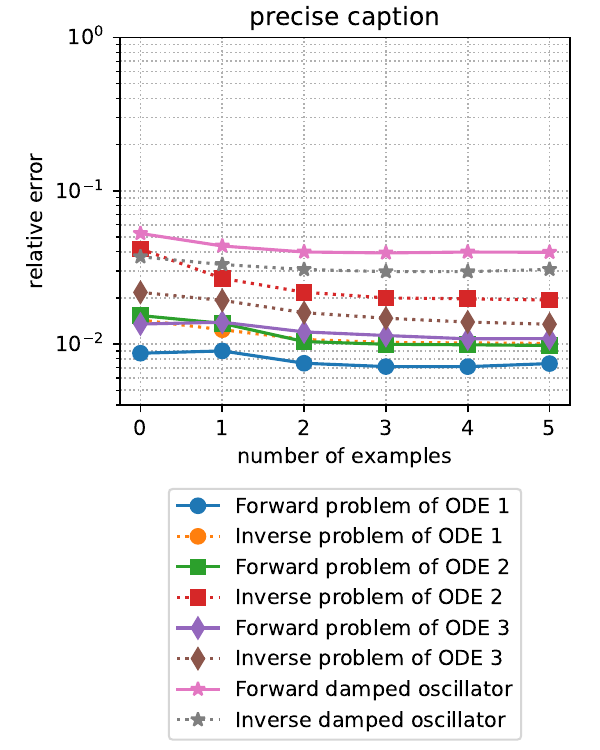}
        \includegraphics[width=\linewidth]{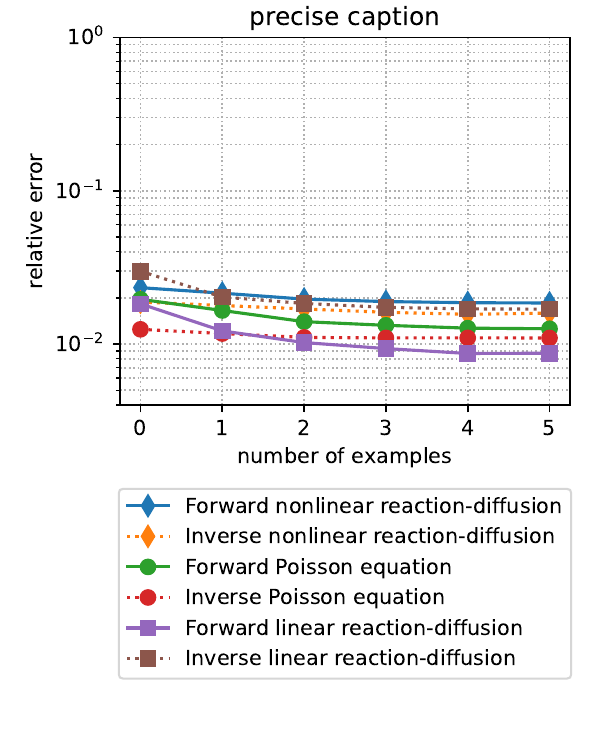}
        \vskip -0.25in
        \includegraphics[width=\linewidth]{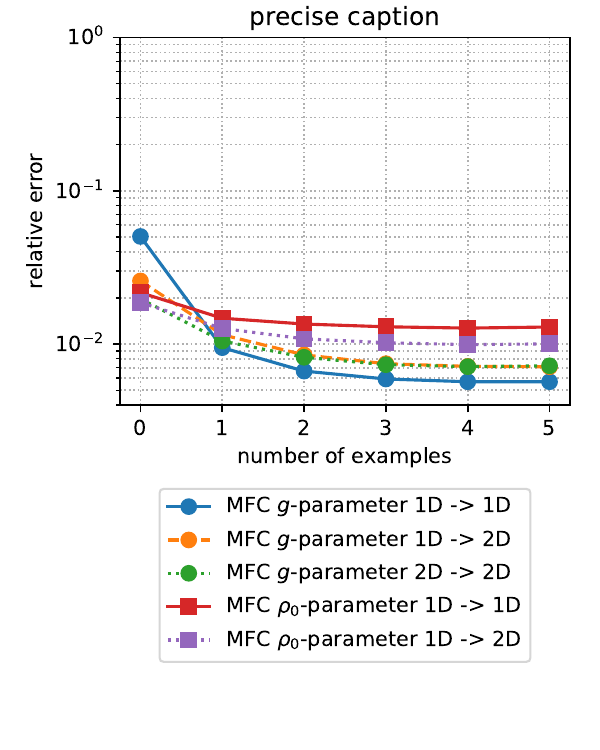}
        \vskip -0.3in
        \caption{}
        \label{fig:caption_error_sub3}
    \end{subfigure}%
    
    \caption{Relative testing error for cases from zero-shot to five-shot learning, for each type of problem. (a) Testing without captions. (b) Testing with vague captions. (c) Testing with precise captions.}
    \label{fig:caption_error}
\end{figure}

We show the performance of the fine-tuned GPT-2 in~\cref{sec:exp_multi} on each specific problem type in Figure~\ref{fig:caption_error}.

\end{document}